\documentclass[runningheads]{llncs}
\usepackage{graphicx}
\usepackage{pgfplots}
\usepackage{subcaption}
\usepackage{mathtools}
\usepackage{hyperref} 
\usepackage{multirow,multicol}
\usepackage{amssymb}
\usepackage{arydshln} 
\usepackage{algorithm,algorithmic}

\pgfplotsset{compat=1.17}
\begin{document}
%
\title {Better Self-training for Image Classification through Self-supervision}
\titlerunning{Better Self-training for Image Classification through Self-supervision}

\author { Attaullah Sahito\and Eibe Frank\and Bernhard Pfahringer}

\date {Australasian Joint Conference on Artificial Intelligence, Sydney 2021.  } 

\authorrunning{Attaullah et al.} 
\tocauthor{Attaullah Sahito, Eibe Frank, and Bernhard Pfahringer
}

\institute{Department of Computer Science, University of Waikato, Hamilton, New Zealand\\
\email{a19@students.waikato.ac.nz}, \email{\{eibe,bernhard\}@waikato.ac.nz}}

\maketitle              
\begin{abstract}
Self-training is a simple semi-supervised learning approach: 
Unlabelled examples that attract high-confidence predictions are labelled with their predictions and added to the training set, with this process being repeated multiple times. Recently, self-supervision---learning without manual supervision by solving an automatically-generated pretext task---has gained prominence in deep learning.
This paper investigates three different ways of incorporating self-supervision into self-training to improve accuracy in image classification: 
self-supervision as pretraining only, self-supervision performed exclusively in the first iteration of self-training, and self-supervision added to every iteration of self-training. 
Empirical results on the SVHN, CIFAR-10, and PlantVillage datasets, using both training from scratch, and Imagenet-pretrained weights, show that applying self-supervision only in the first iteration of self-training can greatly improve accuracy, for a modest increase in computation time. 

\keywords{Self-supervised learning \and Self-training   \and Rotational loss.}
\end{abstract}

\section{Introduction}
Contemporary machine learning based on deep neural networks achieves state-of-the-art results on many tasks, including visual understanding, language modelling, and speech recognition, but often requires a large amount of human-labelled data for such performance. Collecting large amounts of labelled data is expensive and time-consuming. Hence, there is significant interest in algorithms that can harness the benefits of unlabelled data---which is usually plentiful---in addition to the scarce labelled data. Semi-supervised learning (SSL) algorithms are designed for this scenario. Their input comprises
labelled examples $L=\{(x_1,y_1),(x_2,y_2),...,(x_{|L|},y_{|L|})\}$ and unlabelled examples $U= \{{x^{'}_1},{x^{'}_2},...,{x^{'}_{|U|}} \}$, where $x_i, x^{'}_{j} \in X$, with $i= 1,2,...,|L|$  and $j= 1,2,...,|U|$, and $y_i$ are the labels of $x_i$, with $y_i \in \{1,2,3,...,c\}$ and $c$ being the number of classes.   

Self-training is a simple and intuitive generic semi-supervised learning approach. First, a classifier is  trained on labelled examples. Then, the trained model is used to predict labels for unlabelled examples---these labels are called ``pseudo-labels" as they are not actual ground-truth labels in the original data.  Finally, the unlabelled data with its pseudo-labels are merged with the initially labelled data, and the model is retrained on the merged data. The prediction-retraining loop is normally iterated multiple times. Early work on self-training, for word sense disambiguation in text documents, can be found in~\cite{yarowsky1995unsupervised}. 

In contrast, self-supervised learning~\cite{jing2020self} is a fairly recent development in the literature on deep learning. Self-supervised learning is a form of unsupervised learning that works by creating an artificial supervised learning problem based on unlabelled data---a so-called ``pretext" or ``auxiliary" task---for instance, detecting whether an image has been rotated or not. 

In this paper, we empirically compare three different ways of integrating self-supervised learning with self-training. 
In addition to testing them in a setting with random initial weights and training with cross-entropy loss, we also consider the effect of metric learning losses and transfer learning, as they have shown promise in self-training~\cite{oquab2014learning,attaullah2019ssl,sahito20transfer}.

\section{Background}
Semi-supervised learning approaches have been developed since the 1970s~\cite{mclachlan1975iterative}, and self-training, in particular, has continued to attract interest. In the deep learning setting, PseudoLabel~\cite{lee2013pseudo} is an instance of the most basic and straightforward self-training approach, where a neural network model is iteratively trained on initially labelled and pseudo-labelled examples. Although self-training is an appealing and widely-used process, it is worth noting that Arazo et al.~\cite{arazo2020pseudo} found that it can overfit incorrect pseudo-labels, resulting in confirmation bias. Using MixUp~\cite{zhang2018mixup} and enforcing a minimum number of initially labelled examples to be included in each mini-batch was proposed to reduce this bias.

In contrast, self-supervised learning is a more recent development. It is an unsupervised learning method that trains a model  
in a supervised fashion on an artificial supervised learning task generated without any human labelling effort.
The following papers all define such tasks for image recognition:
\begin{itemize}
    \item ExemplarCNN~\cite{dosovitskiy2015discriminative}: $N$ different classes are generated by applying transformations on random image patches, and the network is trained to predict the correct class of a given patch.
    
    \item RotNet~\cite{gidaris2018unsupervised}: Geometric transformations such as rotations by 0, 90, 180, and 270 degrees are applied to an image, and the neural network is trained to predict the rotation applied to an image.
    
    \item Jigsaw Puzzles~\cite{noroozi2016unsupervised}: For a given image, nine patches are generated, and the network is trained to predict the correct permutation order of the patches for that image. 
    
    \item Contrastive learning: The network is trained to differentiate between positive and negative samples, where an image and its augmentations are the positive examples, and all other images are the negative ones~\cite{chen2020simple,chenKSNH20big}. 
\end{itemize}

\section{Self-training using Self-supervised Learning}
Three ways of injecting self-supervision into self-training are proposed below. 
The specific self-supervision task we use in our experiments is to learn to predict one of six possible geometrical transformations: rotations by 0, 90, 180, or 270 degrees respectively, and horizontal or vertical flips. To investigate the effect of the underlying basic loss function in the different variants of self-training algorithms we consider, we evaluate both cross-entropy loss and triplet loss~\cite{schroff2015facenet}. When selecting the most confidently predicted examples in self-training, the highest predicted probability is used as confidence when cross-entropy is applied. Triplet loss, being a metric loss, needs a more complex setup based on a nearest-neighbour classifier applied to embeddings generated by the penultimate network layer. The inverse of the distance to the nearest neighbour is used as  confidence.


\label{section:combine-training}
\begin{algorithm}[t]
\caption{Self-training using Combined Training.}
	\label{alg:combined}
	\begin{algorithmic}[1]
		\STATE\textbf{Input:} Labelled examples ($x_L,y_L$), unlabelled examples $x_U$,  neural network $f_\theta $ with parameters $\theta$, weight of self-supervised loss $\lambda_u$, and  selection percentage $p$.
		\FOR {each meta-iteration}
		\FOR {each epoch over ${x_U \cup x_L}$ } 
		\STATE $b_L$ = sample$(x_L,y_L)$  
		\STATE $b_U$ = geometric-transform(sample$(x_U \cup x_L ))$ 
		\FOR { each mini-batch }
		\STATE $\mathcal{L} = \mathcal{L}_{SUPER}(f_\theta({b_L}))  + \lambda_u \mathcal{L}_{SELF} (f_\theta({b_U}))$
        \STATE $\theta =  \theta -   \nabla_\theta \mathcal{L}	$
        \ENDFOR
		\ENDFOR

		\STATE $labels_U,dist_U =$ assign\_labels$(f_\theta,x_U,x_L,y_L)$
		\STATE $x_{new},y_{new} =$ select\_top$(dist_U,labels_U,p)$ 
		\STATE $x_L,y_L =$  concat$((x_L,y_L),(x_{new},y_{new}))$
		\STATE $x_U =$ delete\_from$(x_U, x_{new})$
		\ENDFOR
	\end{algorithmic}
	\end{algorithm}

\subsection{Combined Training}
\label{section:combined_training}
Combined training (CT) is the most expensive approach we consider: whenever a model is trained in self-training, we jointly train it on both the primary classification task as well as the pretext task, using a combination of self-supervised loss and supervised loss. 
The supervised and self-supervised branches of the model share the same core convolutional neural network, with only the final layers being different. 
The loss is a weighted sum of supervised and self-supervised loss:
\begin{equation}
\label{eqn:combined-loss}
\mathcal{L} = \mathcal{L}_{SUPER}(x_L,y_L)  + \lambda_u \mathcal{L}_{SELF} (x_U,y_U), 
\end{equation}

\noindent where the  hyperparameter $\lambda_u \in [0,1]$ controls the weight of self-supervised loss.

Algorithm~\ref{alg:combined} shows pseudocode for each self-training iteration using combined training (CT). For each epoch, all available training examples, i.e., labelled and unlabelled examples, are employed for self-supervised loss estimation, whereas only the labelled and pseudo-labelled examples are used for supervised loss estimation. 
The network model is used to assign labels to unlabelled examples, and the top $p\%$ most confidently predicted pseudo-labelled examples are selected and merged with the labelled examples for retraining.


\subsection{Self-supervised Pretraining}
\label{section:pretrain}
Pretraining is a common strategy in deep learning: rather than starting with randomly initialised network parameters, we start with parameters obtained by training on a related task. The motivation for pretraining is that the model will learn features that may be useful for the target task. This can be of great importance in semi-supervised learning, where a few labelled and many unlabelled examples are available. It has also been shown that unsupervised pretraining helps neural networks achieve better generalisation~\cite{erhan2010does}. Motivated by this, we consider self-supervision for pretraining on unlabelled examples followed by standard self-training.

Algorithm~\ref{alg:pretrain} presents the training procedure for the proposed self-training approach using self-supervised pretraining (SS-Pretrain). At the start of training, the network model is pretrained on all training examples, i.e., labelled and unlabelled examples using self-supervised loss.\footnote{Note that the original labels are ignored in this step.} Following that, simple self-training is applied using supervised loss only on labelled and pseudo-labelled examples. 

\begin{algorithm}[t]
\caption{Self-training using Self-supervised Pretraining.}
	\label{alg:pretrain}
	\begin{algorithmic}[1]
		\STATE\textbf{Input:} Labelled examples ($x_L,y_L$), unlabelled examples $x_U$,  neural network $f_\theta $ with parameters $\theta$, and  selection percentage $p$
		\FOR {each epoch over $x_U$}
		\STATE $b_U$ = geometric-transform(sample$(x_U \cup x_L ))$
		\FOR { each mini-batch in $x_U \cup x_L$ }
		\STATE $\mathcal{L} = \mathcal{L}_{SELF} (f_\theta({b_U}))$
        \STATE $\theta =  \theta -   \nabla_\theta \mathcal{L}	$
		\ENDFOR
		\ENDFOR
		\FOR {each meta-iteration}
		\STATE $f_\theta =$ train\_network$(x_L,y_L)$ 
		\STATE $labels_U,dist_U =$ assign\_labels$(f_\theta,x_U,x_L,y_L)$
		\STATE $x_{new},y_{new} =$ select\_top$(dist_U,labels_U,p)$ 
		\STATE $x_L,y_L =$  concat$((x_L,y_L),(x_{new},y_{new}))$
		\STATE $x_U =$ delete\_from$(x_U, x_{new})$
		\ENDFOR
	\end{algorithmic}
\end{algorithm}

\subsection{Self-training using Single-Step Combined Training}
\label{section:stssc}
In combined training (see Section~\ref{section:combine-training}), self-supervised training is applied in each epoch during all self-training iterations. 
This is rather expensive. Moreover, it is conceivable that self-training is no longer beneficial once the model has been sufficiently adapted to the primary task: it may destroy features that are finely tuned to the primary task at hand. Therefore, we also investigate using self-supervision only in the first iteration of self-training.
Algorithm~\ref{alg:stssc} shows this self-training approach using single-step combined training (STSSC). 

\begin{algorithm}[t]
\caption{Self-training using Single-step Combined Training.}
	\label{alg:stssc}
	\begin{algorithmic}[1]
		\STATE\textbf{Input:} Labelled examples ($x_L,y_L$), unlabelled examples $x_U$,  neural network $f_\theta $ with parameters $\theta$, weight of self-supervised loss $\lambda_u$, and  selection percentage $p$.
		
		\FOR {each epoch over $x_U$}
		\STATE $b_L$ = sample$(x_L,y_L)$
		\STATE $b_U$ = geometric-transform(sample$(x_U \cup x_L ))$
		\FOR { each mini-batch in $x_U$ }
		\STATE $\mathcal{L} = \mathcal{L}_{SUPER}(f_\theta({b_L}))  + \lambda_u \mathcal{L}_{SELF} (f_\theta({b_U}))$
        \STATE $\theta =  \theta -   \nabla_\theta \mathcal{L}	$
		\ENDFOR
		\ENDFOR
		
		\FOR { each meta-iteration}
		\STATE $f_\theta =$ train\_network$(x_L,y_L)$ 
		\STATE $labels_U,dist_U =$ assign\_labels$(f_\theta,x_U,x_L,y_L)$
		\STATE $x_{new},y_{new} =$ select\_top$(dist_U,labels_U,p)$ 
		\STATE $x_L,y_L =$  concat$((x_L,y_L),(x_{new},y_{new}))$
		\STATE $x_U =$ delete\_from$(x_U, x_{new})$
		\ENDFOR
	\end{algorithmic}
	
\end{algorithm}

\section{Empirical Evaluation}
For evaluating the effect of self-supervised learning in self-training, we conduct an extensive comparison using standard benchmark datasets.
We always report three different test accuracies: (a) after training on the initially labelled examples, (b) after full self-training, and (c) after training on all-labelled training examples. 
Results obtained for (a) and (c) act as empirical lower and upper bounds for self-training.
All results are averaged over three runs with a random selection of initially labelled examples. 

\subsection{Datasets}
We use three benchmark datasets: Street View House Numbers (SVHN)~\cite{netzer2011reading}, CIFAR-10~\cite{krizhevsky2010convolutional}, and PlantVillage~\cite{hughes2015open}. 
PlantVillage images have been resized to 96 by 96, 64 by 64, and 32 by 32, respectively, to create three datasets.

Following standard semi-supervised learning practice, a small class-balanced subset of labelled examples is chosen randomly for each dataset.
All remaining training examples are used as unlabelled examples. The number of initially labelled examples is 1000 for SVHN, 4000 for CIFAR-10, and 380 for PlantVillage.

\subsection{Convolutional Architecture}
A wide residual network with depth 28 and width 2 (WRN-28-2) is a common choice in semi-supervised classification tasks~\cite{oliver2018realistic}. 
In our experiments, we add a fully connected layer at the end of the WRN-28-2 model to produce 64-dimensional embeddings. For all results, to obtain insight into the effect of standard ImageNet pretraining on the outcome of semi-supervised learning, we report test accuracy obtained by running our algorithms by starting with a) randomly initialised network weights and b) ImageNet pretrained weights. For pretraining, downsampled ImageNet~\cite{chrabaszcz2017downsampled} images with sizes 32 by 32 are used.

\subsection{Algorithm Configurations}

For SVHN and CIFAR-10, a mini-batch size of 100 is used, while PlantVillage uses a batch size of 64. Adam~\cite{kingma2014adam} is used as an optimiser for updating the network parameters with a learning rate of $10^{-3}$ for randomly initialised weights and $10^{-4}$ for ImageNet pretrained weights. After the first iteration of self-training, the learning rate is further reduced by a factor of $0.1$. Triplet loss is estimated using $l2$-normalised embeddings and a margin value of $1$. A  $1$-nearest-neighbour classifier is employed on the embeddings to compute test accuracy. Self-training is applied for 25 meta-iterations. For SVHN and CIFAR-10,  $p=5\%$ pseudo-labels are selected from unlabelled examples, while PlantVillage uses $p=2\%$. For self-supervised loss, the weight parameter $\lambda_u $ is set to $1$ to avoid tuning it separately for every single dataset\footnote{Source code is available at \url{https://github.com/attaullah/Self-training/blob/master/Self_supervised.md}}.  

\subsection{Results: Detailed Evaluation of the Three Algorithms}

Table~\ref{table:combined-all} shows test accuracy for CIFAR-10, SVHN, and  the three PlantVillage datasets for the WRN-28-2 network using random and pretrained (ImageNet) weights based on combined training (see Section~\ref{section:combine-training}). Cross-entropy loss is used for the self-supervised component of the combined loss in Equation 1, but we consider both cross-entropy loss and triplet loss for the supervised component. 
Values in bold highlight the best test accuracy for each dataset for (a) the empirical lower bound provided by training on initial labels (``N-Labelled"), (b) self-training using combined training, and (c) the upper bound provided by training with all labels (``All-Labelled"), across the two loss function configurations and network initialisations considered. Results for triplet loss are obtained by applying a 1-nearest-neighbour classifier to the learned embeddings.

\begin{table}[t]
    \centering
    \caption{Test accuracy on WRN-28-2 using random and ImageNet pretrained weights for Combined Training (CT).}
    \label{table:combined-all}
    \resizebox{\textwidth}{!}{%
    \begin{tabular}{|l|l|r|r|r|r|r|r|}
    \hline
    &    & \multicolumn{2}{c|}{\textbf{N-Labelled}} & \multicolumn{2}{c|}{\textbf{Self-training}} & \multicolumn{2}{c|}{\textbf{All-Labelled}} \\ \hline
    
    & \textbf{Weights} & \textbf{Cross-entropy}  & \multicolumn{1}{l|}{\textbf{Triplet}} & \textbf{Cross-entropy}    & \multicolumn{1}{l|}{\textbf{Triplet}}   & \textbf{Cross-entropy}   & \multicolumn{1}{l|}{\textbf{Triplet}}  \\ \hline
    
    \multirow{2}{*}{\textbf{SVHN}}  & Random  & 
     $ 86.37 \pm 3.57 $ & $ 89.38 \pm 1.69 $ & $ 88.10 \pm 3.02 $ & $ 79.47 \pm 0.82 $& $ \mathbf{95.74 \pm 0.02} $ & $ 95.07 \pm 0.15 $ \\  
    & ImageNet & $ \mathbf{89.62 \pm 1.22} $ & $ 88.98 \pm 2.14 $ & $ \mathbf{90.83 \pm 1.67} $ & $ 81.96 \pm 3.24 $& $95.16 \pm 0.16 $& $ 94.65 \pm 0.35 $ \\ \hline
       
    \multirow{2}{*}{\textbf{CIFAR-10}}          & Random           & 
    $ 80.67 \pm 2.95 $ & $ 81.25 \pm 1.77 $ & $ 86.04  \pm 0.60 $ & $ 81.51 \pm 0.73 $& $ 91.38 \pm 0.39 $& $ 86.97 \pm 3.47 $ \\  
    & ImageNet     & $ 87.00 \pm 1.97 $ & $ \mathbf{90.04 \pm 1.22} $ & $ \mathbf{89.19 \pm 0.65} $ & $ 89.11 \pm 2.33 $& $ 93.47 \pm 0.08 $& $ \mathbf{93.92 \pm 0.09 }$ \\ \hline
    
    \multirow{2}{*}{\textbf{PLANT32}} & Random           & 
    $  67.46  \pm 3.39 $ & $ 55.27 \pm 3.77  $ & $ 72.91 \pm 3.68 $ & $ 50.99  \pm 1.21 $& $ 97.97 \pm 0.64 $& $ 91.23 \pm 0.86 $ \\  
    & ImageNet     & $ \mathbf{77.96 \pm 0.58} $ & $ 73.38  \pm 1.28 $ & $ \mathbf{78.18 \pm 0.09} $ & $ 66.67   \pm 0.60 $& $ \mathbf{98.38 \pm 0.08} $ & $ 96.58  \pm 0.48 $  \\ \hline
     
    \multirow{2}{*}{\textbf{PLANT64}} & Random           & 
     $ 72.52 \pm 3.86 $ & $ 63.86  \pm 0.94 $ & $ 83.52 \pm 1.84  $ & $ 59.84  \pm 0.38 $ & $ 98.52 \pm 0.34 $& $ 91.99 \pm 0.30 $ \\  
    & ImageNet     &  $ \mathbf{78.39 \pm 0.87} $ & $ 76.65 \pm 1.06 $ & $ \mathbf{84.98 \pm 0.68} $ & $ 69.26  \pm 6.11 $ & $ \mathbf{99.08 \pm 0.10} $ & $ 94.67   \pm 0.52 $ \\ \hline
    
    \multirow{2}{*}{\textbf{PLANT96}} & Random           & 
    $ 75.12  \pm 3.91 $ & $ 65.94  \pm 1.43 $ & $ 79.94  \pm 1.75 $ & $ 44.55  \pm 1.70 $ & $ 98.08 \pm 0.65 $& $ 89.47  \pm 0.99 $ \\  
    & ImageNet     & $ \mathbf{79.07 \pm 0.68} $ & $  79.04  \pm 0.67 $ & $ \mathbf{84.48  \pm 1.30} $ & $ 76.26  \pm 1.40 $ & $ \mathbf{99.15 \pm 0.12} $ & $ 95.60 \pm 1.27 $ \\  \hline
    \end{tabular}%
    }
    
\end{table}

Excluding the results obtained for the lower and upper bounds based on purely supervised training on the CIFAR-10 data with ImageNet pretraining, it is clear that there is no benefit to be had by using metric learning with triplet loss. The results also show that self-training using cross-entropy provides consistent improvements compared to the lower bound, but there is a large gap compared to the results for the upper bound.
We can also see that it is generally highly beneficial to use ImageNet weights instead of randomly initialised weights, regardless of dataset, loss function, and learning algorithm. The few observed drops in accuracy for the SVHN data are very small.

Table~\ref{table:pretrain-table} shows test accuracy for CIFAR-10, SVHN, and PlantVillage32, 64, 96 on WRN-28-2 using random and ImageNet weights after applying SS-Pretrain (see Section~\ref{section:pretrain}). 
To apply pretraining based on self-supervision, we run it on all training examples---those that exhibit a ground-truth label as well as those missing one---for 120 epochs.
After that, we run 200 epochs of fine-tuning on only the labelled training data, followed by standard self-training with 25 meta-iterations, using either cross-entropy or triplet loss. To provide a fair comparison, we also apply self-supervision in the same manner to obtain initial weights for the upper and lower bounds obtained by purely supervised training.

\begin{table}[t]
    \centering
    \caption{Test accuracy on WRN-28-2 using random and ImageNet pretrained weights for SS-Pretrain.}
    \label{table:pretrain-table}
    \resizebox{\textwidth}{!}{%
    \begin{tabular}{|l|l|r|r|r|r|r|r|}
    \hline
    &    & \multicolumn{2}{c|}{\textbf{N-Labelled}} & \multicolumn{2}{c|}{\textbf{Self-training}} & \multicolumn{2}{c|}{\textbf{All-Labelled}} \\ \hline
    
    & \textbf{Weights} & \textbf{Cross-entropy}  & \multicolumn{1}{l|}{\textbf{Triplet}} & \textbf{Cross-entropy}    & \multicolumn{1}{l|}{\textbf{Triplet}}   & \textbf{Cross-entropy}   & \multicolumn{1}{l|}{\textbf{Triplet}}  \\ \hline
    
    \multirow{2}{*}{\textbf{SVHN}}  & Random  & 
    $ 89.51 \pm 0.74  $ & $ 86.48 \pm 2.00 $ & $ \mathbf{89.72 \pm 0.57}   $ & $ 83.73 \pm 2.46 $ & $ 95.63 \pm 0.46 $& $ 95.05 \pm 0.06 $ \\  
    & ImageNet & $ 89.90 \pm 0.40 $ & $ \mathbf{90.08 \pm 0.60} $ & $ 89.67 \pm 0.10 $ & $ 86.05 \pm 0.44 $& $ \mathbf{95.90 \pm 0.05} $& $ 95.30 \pm 0.16 $  \\ \hline
       
    \multirow{2}{*}{\textbf{CIFAR-10}}          & Random           & 
    $  72.02 \pm 0.50 $ & $ 75.76 \pm 2.11 $ & $ 85.50 \pm 0.41  $ & $ 75.51 \pm 1.88 $& $ 90.32 \pm 0.94 $& $ 89.11 \pm 0.68 $ \\  
    & ImageNet & $ 84.83 \pm 1.99 $ & $ \mathbf{ 86.98 \pm 0.84 } $ & $ \mathbf{ 91.06 \pm 0.47 } $ & $ 90.18 \pm 0.87 $& $ \mathbf{93.48 \pm 0.09} $ & $ 92.64 \pm 0.70 $ \\ \hline
    
    \multirow{2}{*}{\textbf{PLANT32}} & Random           & 
    $ 69.80 \pm 1.07 $ & $ 72.03 \pm 1.51 $ & $ 68.72 \pm 0.30 $ & $ 65.04 \pm 1.19 $& $ 97.14 \pm 0.68 $& $ 96.64 \pm 0.07 $ \\  
    & ImageNet  & $ 72.38 \pm 1.22 $ & $ \mathbf{ 72.46 \pm 1.07 } $ & $ \mathbf{75.21 \pm 1.34 } $ & $ 69.96 \pm 0.05 $& $ \mathbf{99.24 \pm 0.12} $ & $ 97.77 \pm 0.77 $ \\ \hline
     
    \multirow{2}{*}{\textbf{PLANT64}} & Random           & 
    $ 73.35 \pm 1.76 $ & $ 71.10 \pm 1.16 $ & $ 73.45 \pm 3.32 $ & $ 67.06 \pm 1.42 $ & $ 97.94 \pm 0.48 $& $ 87.76 \pm 0.96 $ \\   
    & ImageNet  & $ 75.64 \pm 0.54 $ & $ \mathbf{ 77.51 \pm 1.05} $ & $ \mathbf{75.50 \pm 0.93} $ & $ 68.37 \pm 1.43$ & $ \mathbf{99.06 \pm 0.27} $ & $ 89.48 \pm 0.56 $  \\ \hline
    
    \multirow{2}{*}{\textbf{PLANT96}} & Random           & 
    $ 73.29 \pm 0.72 $ & $ 70.61 \pm 0.83 $ & $ 75.19 \pm 0.88 $ & $ 67.18 \pm 1.81 $ & $ 98.43 \pm 0.05 $ & $ 90.54 \pm 1.18    $ \\  
    & ImageNet  & $ 78.37 \pm 1.19 $ & $ \mathbf{78.74 \pm 0.22} $ & $ \mathbf{79.11 \pm 1.24} $ & $ 74.97 \pm 0.57 $ & $ \mathbf{99.47 \pm 0.12} $& $ 93.32 \pm 1.35 $ \\  \hline
    \end{tabular}
    }
    
\end{table}

Values in bold highlight the best test accuracy for each dataset obtained by performing self-supervised pretraining for (a) the empirical lower bound provided by training on initial ground-truth labels (``N-Labelled"), (b) self-training, and (c) the upper bound provided by training with all ground-truth labels (``All-Labelled"), across the two loss function configurations and network initialisations considered. Interestingly, metric learning using triplet loss now often achieves better accuracy than applying cross-entropy loss when using only the initially labelled examples. However, cross-entropy consistently achieves higher test accuracy than metric learning when performing self-training and when training using all labelled examples. Comparing self-training to the lower bound, we see improvements when using cross-entropy, but self-training reduces accuracy most of the time when considering metric learning. Networks initialised with ImageNet weights generally perform better than those with randomly initialised weights.

Table~\ref{table:self-vs-simple-table} shows test accuracy of self-training with STSSC (see Section~\ref{section:stssc}) for CIFAR-10, SVHN, and PlantVillage32, 64, 96 on WRN-28-2 using random and ImageNet pretrained weights. The reported results are obtained after one iteration of combined training and a further 24 iterations of regular self-training, using either cross-entropy or triplet loss. Values in bold highlight the best test accuracy for each dataset across the two loss function configurations and network initialisations considered. For the corresponding empirical lower and upper bounds, please consult the results in Table~\ref{table:combined-all}. Self-training using cross-entropy outperforms metric learning for all five datasets for both randomly initialised and ImageNet pretrained weights. Moreover, in the majority of the cases, using ImageNet weights achieves higher test accuracy than using random weights. Notably, for the one dataset where random weights yield higher accuracy, the difference in performance is very small.

\begin{table}[t]
    \centering
    \caption{Self-training test accuracy on WRN-28-2 using random and ImageNet pretrained	weights for STSSC.}
    \label{table:self-vs-simple-table}
    \begin{tabular}{|l|l|r|r|}
    \hline
      & \textbf{Weights} & \textbf{Cross-entropy}  & \multicolumn{1}{l|}{\textbf{Triplet }}   \\ \hline
    
    \multirow{2}{*}{\textbf{SVHN}} & Random & 
     $ \mathbf{95.27 \pm 0.20} $ & $ 83.62 \pm 0.03 $  \\  
    & ImageNet &  $ 95.12 \pm 0.32 $ & $ 83.47 \pm 0.98 $  \\  \hline
    
    \multirow{2}{*}{\textbf{CIFAR-10}} & Random &         
    $ 87.24 \pm 0.67 $ & $ 79.36 \pm 1.02 $  \\  
    & ImageNet & $ \mathbf{91.64 \pm 0.38 } $ & $ 90.75 \pm 1.48 $  \\  \hline
    
    \multirow{2}{*}{\textbf{PLANT32}} & Random & 
    $ 77.39 \pm 1.32 $ & $ 54.44 \pm 3.72 $  \\  
    & ImageNet & $ \mathbf{86.72 \pm 1.10} $ & $ 65.14 \pm 1.85 $  \\  \hline
    
    \multirow{2}{*}{\textbf{PLANT64}} & Random & 
    $ 86.13 \pm 0.46 $ & $ 56.23 \pm 2.01 $  \\ 
    & ImageNet &  $ \mathbf{89.48 \pm 0.37} $ & $ 74.29 \pm 1.27 $  \\ \hline
    
    \multirow{2}{*}{\textbf{PLANT96}} & Random & 
    $ 84.71 \pm 1.89 $ & $ 58.14 \pm 0.82 $  \\  
    & ImageNet & $ \mathbf{89.95 \pm 0.37} $ & $ 74.41 \pm 3.75 $ \\  \hline
    \end{tabular}%
\end{table}

\begin{figure}[t]
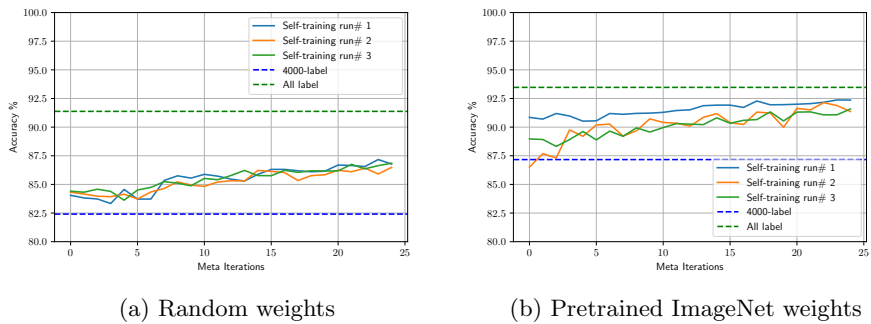

	\centering 
	\begin{subfigure}{0.5\textwidth}
		\centering
		\resizebox{1.\textwidth}{!} {\input{images/selflogs_self-simple.pgf} }
		\caption{Random weights}
		\label{fig:cifar10-random-self-train-only}
	\end{subfigure}\hfil 
	\begin{subfigure}{0.5\textwidth}
		\centering	
		\resizebox{1.\textwidth}{!} {\input{images/selflogs_self-simple-w.pgf} }
		\caption{Pretrained ImageNet weights}
		\label{fig:cifar10-pretrained-selftrain-only}
	\end{subfigure} 
	
	\label{fig:cifar10-selftriplet-random-pretrained}
	\caption{Comparison of self-training using STSSC on WRN-28-2 for CIFAR-10 using cross-entropy loss.}
	\label{fig:cifar10-self-train-only}
\end{figure}

To further illustrate the effect of starting the training process with ImageNet weights rather than random weights, Figure~\ref{fig:cifar10-self-train-only} plots the test accuracy for STSSC over the meta-iterations across three different runs using a) randomly initialised weights and b) ImageNet weights on CIFAR-10 using 4000 labelled examples initially. Cross-entropy loss is used as the loss function throughout. The accuracy curves show definite improvements as more meta-iterations are performed, for both weight initialisations, but the ImageNet version starts from a higher initial accuracy and retains this advantage over the 25 meta-iterations of self-training.

\subsection{Comparing Self-training Approaches}
Table~\ref{table:comparison-all-configs} presents a summary view and comparison of the three approaches to integrating self-supervision into self-training, and, importantly, reports results for self-training without using any self-supervised learning---denoted by the name ``Simple" in the table. 
Cross-entropy loss is used for both supervised and self-supervised loss estimation throughout. The results show that STSSC performs best overall, demonstrating the importance of applying self-supervised learning in a judicious manner when combining it with self-training: applying it in all meta-iterations is generally better than simply using it for pretraining, but among the three algorithm variants considered, it is clearly best to only apply it in the first meta-iteration.
Compared to the simple self-training baseline without self-supervision, STSSC performs substantially better when starting with randomly initialised weights. However, while the performance also generally improves when using ImageNet weights, and substantially so for the SVHN data, STSSC only wins in three cases, with the simple approach performing best for the two smaller-sized versions of the PlantVillage datasets.

\begin{table}[t]
    \centering
    \caption{Test accuracy after self-training on WRN-28-2 using various self-supervised settings.}
    \label{table:comparison-all-configs}
    \resizebox{\textwidth}{!}{%
    \begin{tabular}{|l|r|r|r|r|r|}
    \hline
    & \multicolumn{1}{l|}{\textbf{SVHN}}  & \multicolumn{1}{l|}{\textbf{CIFAR-10}} & \multicolumn{1}{l|}{\textbf{PLANT32}} & \multicolumn{1}{l|}{\textbf{PLANT64}} & \multicolumn{1}{l|}{\textbf{PLANT96}} \\ \hline
    
    \multicolumn{6}{|l|}{ \textbf{Random weights}}  \\ \hline
    
    \textbf{Simple} &  $ 94.28 \pm 0.45 $     & $ 83.05 \pm 0.99 $     & $ 68.24 \pm 2.44 $     & $ 69.36 \pm 1.80 $     & $ 76.09 \pm 2.35 $   \\ \hdashline
    
    \textbf{CT} & $ 88.10 \pm 3.02 $     & $ 86.04 \pm 0.60 $     & $ 72.91 \pm 3.68 $     & $ 83.52 \pm 1.84 $     & $ 79.24 \pm 1.45 $    \\ 
    
    \textbf{SS-Pretrain}  & $ 89.72 \pm 0.57 $     & $ 85.50 \pm 0.41 $     & $ 68.72 \pm 0.30 $     & $ 73.45 \pm 3.32 $     & $ 75.19 \pm 0.88 $    \\
    
    \textbf{STSSC} & $ \mathbf{95.27 \pm 0.20} $     & $ \mathbf{87.08 \pm 0.66} $     & $ \mathbf{77.39 \pm 1.32} $     & $ \mathbf{ 86.13 \pm 0.46} $     & $ \mathbf{84.71 \pm 1.89} $    \\    \hline
    
    \multicolumn{6}{|l|}{ \textbf{ImageNet weights}}   \\ \hline
    
    \textbf{Simple} & $ 90.84 \pm 0.48 $     & $ 91.62 \pm 0.58  $  & $ \mathbf{87.48 \pm 0.84} $     & $ \mathbf{92.60 \pm 1.03} $  & $ 89.65 \pm 0.88 $    \\ \hdashline
    
    \textbf{CT} & $ 90.83 \pm 1.67 $     & $ 89.19 \pm 0.65 $     & $ 78.18 \pm 0.09 $     & $ 84.60 \pm 0.77 $     & $ 84.48 \pm 1.30 $    \\
    
    \textbf{SS-Pretrain} & $ 89.67 \pm 0.10 $     & $ 91.06 \pm 0.47 $     & $ 75.21 \pm 1.34 $     & $ 75.50 \pm 0.93 $     & $ 79.11 \pm 1.24 $    \\
    
    \textbf{STSSC}& $ \mathbf{95.12 \pm 0.32} $     & $ \mathbf{91.65 \pm 0.43} $     & $ 86.72 \pm 1.10 $     & $ 89.48 \pm 0.37 $     & $ \mathbf{89.95 \pm 0.37} $    \\     \hline
    
    \end{tabular}%
    }
\end{table}

\begin{figure} [t]
	\centering 
		\resizebox{0.5\textwidth}{!} {\input{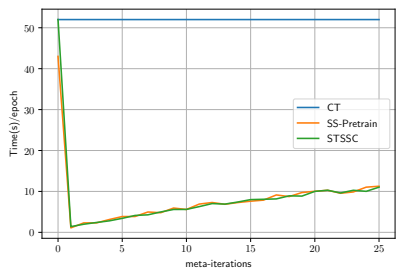} }
    \caption{Time spent on each epoch over meta-iterations of self-training approaches using self-supervised learning.}
 \label{fig:train-time}
\end{figure}

Figure~\ref{fig:train-time} compares the time needed for each epoch over the self-training meta-iterations for all three self-supervision-based self-training approaches, i.e., CT, SS-Pretrain, and STSSC. To estimate time, WRN-28-2 is trained on CIFAR-10 with 4000-labelled examples initially for 25 meta-iterations of self-training. Time spent by each epoch for CT is constant and higher than for SS-Pretrain and STSSC. After the first iteration, SS-Pretrain and STSSC need very little time compared to CT.

\subsection{Visualising Embeddings}
\begin{figure}[t]
	\centering 
	\begin{subfigure}{0.5\textwidth}
		\centering
		\includegraphics[scale=0.167] {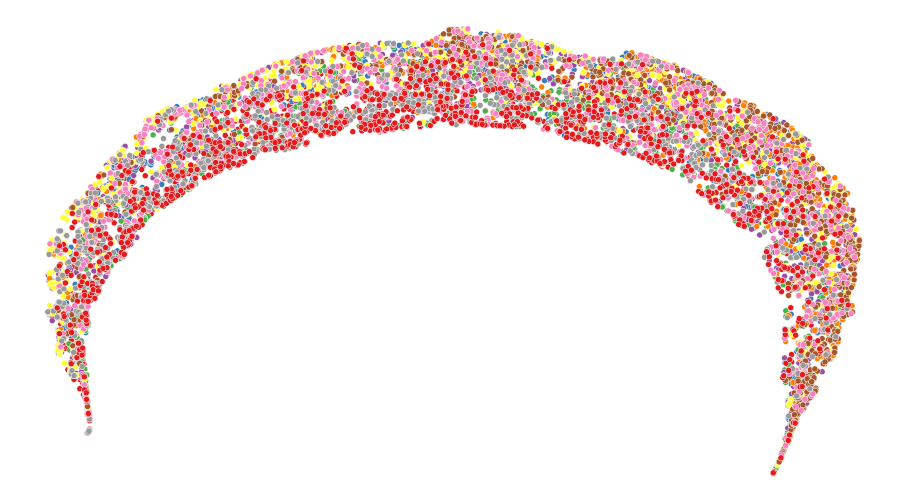}
		\caption{ Randomly initialised weights}
		\label{fig:cifar10-wrn-random-self}
	\end{subfigure}\hfil 
	\begin{subfigure}{0.5\textwidth}
		\includegraphics[scale=0.167] {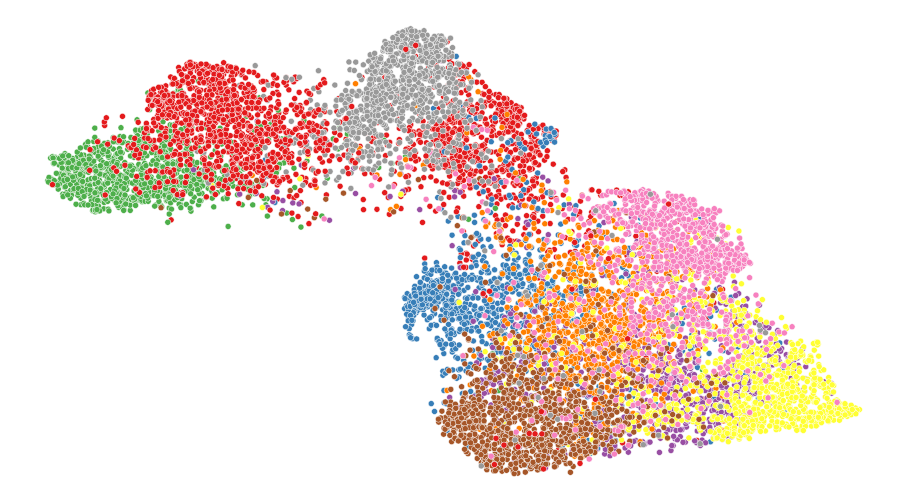}
		\caption{ ImageNet weights}
		\label{fig:cifar10-wrn-w-self}
	\end{subfigure}
	
	\medskip
	
	\begin{subfigure}{0.5\textwidth}
		\centering
		\includegraphics[scale=0.167] {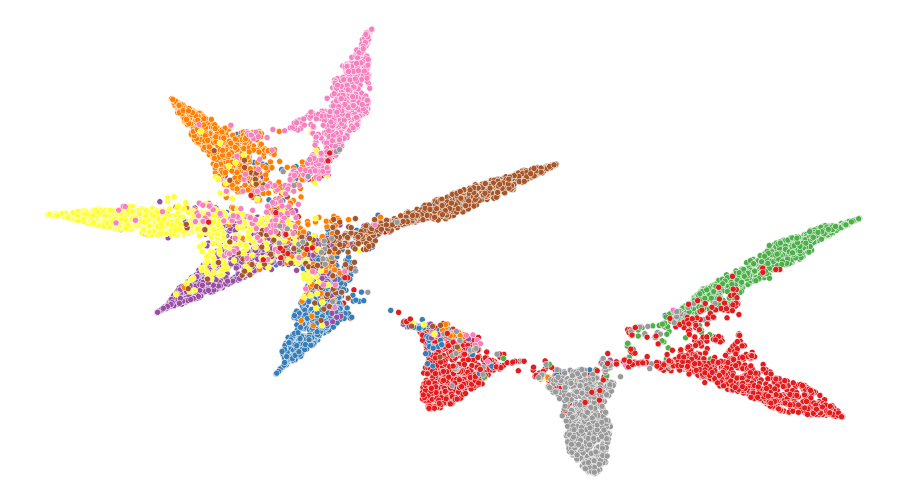}
		\caption{After one iteration of CT/STSSC}
		\label{fig:cifar10-wrn-finw-nw}
	\end{subfigure}\hfil 
	\begin{subfigure}{0.5\textwidth}
		\includegraphics[scale=0.167] {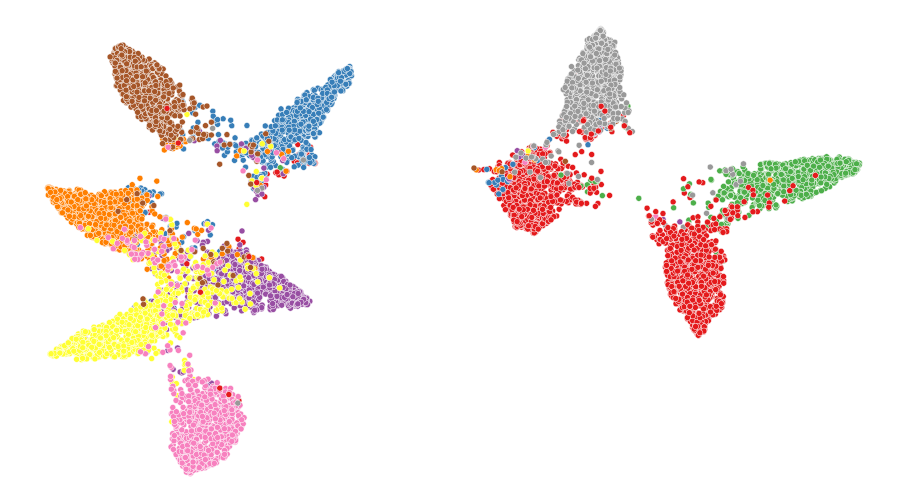}
		\caption{After one iteration of CT/STSSC}
		\label{fig:cifar10-wrn-fine-w}
	\end{subfigure}
	
	\medskip
	
	\begin{subfigure}{0.5\textwidth}
		\centering
		\includegraphics[scale=0.167] {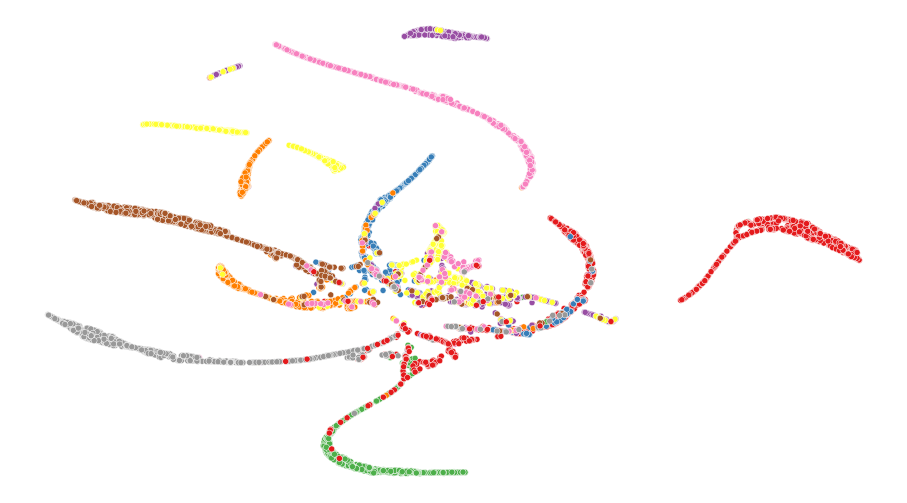}
		\caption{After full self-training with STSSC }
		\label{fig:cifar10-wrn-meta-nw}
	\end{subfigure}\hfil 
	\begin{subfigure}{0.5\textwidth}
	    \includegraphics[scale=0.167] {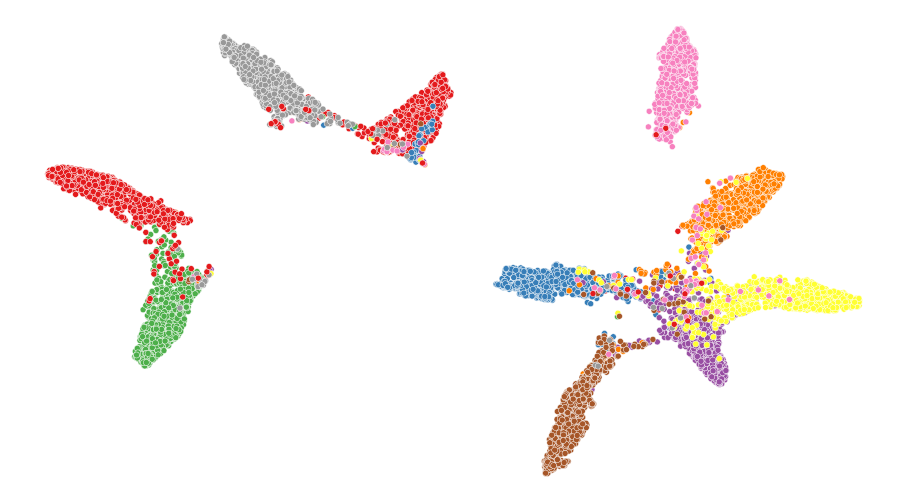}
		\caption{After full self-training with STSSC }
		\label{fig:cifar10-wrn-meta-w}
	\end{subfigure}
	
	\medskip
	
	\begin{subfigure}{\textwidth}
		\centering
        \resizebox{0.95\textwidth}{!} {\input{images/legend.pgf}}
		\label{fig:legend}
	\end{subfigure} 
	
\caption{UMAP visualisation of CIFAR10 test set embeddings obtained from WRN-28-2 using random weights [left: (a),(c),(e)] and ImageNet weights [right: (b),(d),(f)] by applying STSSC training.}
	\label{fig:STSSC-cifar10-wrn-randomvspretrain}
\end{figure}

Embeddings generated for CIFAR-10 by STSSC are visualised in  Figure~\ref{fig:STSSC-cifar10-wrn-randomvspretrain},
using UMAP~\cite{mcinnes2018umap} to produce two-dimensional projections.
The embeddings are obtained at the start of training: (a) and (b); after one iteration of CT: (c) and (d); and after 25 meta-iterations of self-training: (e) and (f). Configurations (a), (c), and (e) use randomly initialised weights, (b), (d), and (f) use ImageNet weights. Self-training clearly generates well-separated clusters.

\section{Conclusion}
In this paper, we have shown that self-supervision can substantially improve self-training for image classification problems. Self-supervision was applied in three different ways.
The best performance was achieved by using self-supervision only in the first iteration of self-training, with cross-entropy loss and initial weights obtained from ImageNet.
There are a number of directions for future research, including more complex scenarios such as self-supervised pretraining followed by a limited number of combined training meta-iterations, followed by further regular self-training. Alternatively, one could consider the use of different and potentially more challenging self-supervision tasks. 

 
\bibliographystyle{splncs04}
\bibliography{Bibliography}

\end{document}